\documentclass[10pt,twocolumn]{article}

\usepackage[a4paper,margin=1.72cm,columnsep=0.62cm]{geometry}
\usepackage{fontspec}
\newfontfamily\cjkfont{FandolSong-Regular.otf}[
  BoldFont=FandolSong-Bold.otf,
  ItalicFont=FandolSong-Regular.otf,
  BoldItalicFont=FandolSong-Bold.otf
]
\usepackage{microtype}
\usepackage{graphicx}
\usepackage{booktabs}
\usepackage{tabularx}
\usepackage{array}
\usepackage{enumitem}
\usepackage{amsmath,amssymb}
\usepackage{tikz}
\usetikzlibrary{arrows.meta,positioning,fit,shapes.geometric}
\usepackage{xcolor}
\definecolor{ink}{HTML}{263238}
\definecolor{blueaccent}{HTML}{2F6690}
\definecolor{orangeaccent}{HTML}{C66A2B}
\definecolor{lightblue}{HTML}{EAF2F8}
\definecolor{lightorange}{HTML}{FBF0E8}
\usepackage{caption}
\usepackage{float}
\usepackage{url}
\usepackage[hidelinks]{hyperref}
\usepackage[numbers,sort&compress]{natbib}

\hypersetup{pdftitle={Knowing the Rules, Applying the Rules: Evaluating Language Models on Traditional Chinese Bazi},pdfauthor={Jiulin Li and Ping Huang}}
\setlist{nosep,leftmargin=*}

\newcommand{\code}[1]{\texttt{#1}}
\newcommand{\tightfig}[2][\linewidth]{\includegraphics[width=#1]{#2}}
\newcolumntype{Y}{>{\raggedright\arraybackslash}X}
\newcolumntype{R}{>{\raggedleft\arraybackslash}p{1.25cm}}

\title{Knowing the Rules, Applying the Rules:\\Evaluating Language Models on Traditional Chinese Bazi}
\author{Jiulin Li\textsuperscript{1,*}\qquad Ping Huang\textsuperscript{2}\\[4pt]
\small\textsuperscript{1}Beijing Liuyi Guanhua Technology Co., Ltd.\\
\small\textsuperscript{2}State Key Laboratory of General Artificial Intelligence, BIGAI\\[3pt]
\small *Corresponding author: \href{mailto:lijiulin@18trees.com}{lijiulin@18trees.com}}
\date{September 2026}

\begin{document}
\raggedbottom
\maketitle
\begin{abstract}
Knowing domain rules does not guarantee applying them to a case. We study this distinction in traditional Chinese Bazi through 3,000 Chinese multiple-choice questions spanning 14 Theory and 11 Case categories. Six endpoint systems are evaluated, with primary results reported on a 2,492-item model-informed refinement. Theory accuracy exceeds Case accuracy for every system, and gaps of 16.60--29.56 percentage points remain when invalid responses are excluded. The contrast is more specific than a general case-reasoning deficit. Across six systems, Twelve Stages and Nayin reach mean accuracies of 89.10\% and 88.62\%, while Shensha Basics reaches 75.96\%. Within Case, Luck Pillars averages 84.62\%, but Career and Family Relations average only 36.98\% and 38.19\%. Overall rankings also conceal different category strengths. On the original 3,000 items, paired DeepSeek native/disabled comparisons associate native configurations with Theory gains of 6.53 and 12.20 points for Flash and Pro, respectively; Case changes are -3.67 and +1.27 points. These are provider-configuration associations, not isolated causal effects of reasoning. The results motivate task-specific evaluation of cultural-domain applications rather than reliance on aggregate knowledge scores. The benchmark measures agreement with a model-generated, model-verified answer key, not real-world predictive validity. Final-set results are post-selection descriptions, and incomplete provenance and expert validation constrain their interpretation.
\end{abstract}

% First-page single-column benchmark anatomy. Counts and bar geometry are deterministic.
% Generated imagery is a conceptual visual anchor only; it contains no data.
\begin{figure}[t]
\centering
\begin{tikzpicture}[x=1mm,y=-1mm,
  every node/.style={font=\sffamily\fontsize{8}{9.3}\selectfont,text=ink,inner sep=0pt},
  heading/.style={font=\sffamily\bfseries\fontsize{9}{10.5}\selectfont},
  small/.style={font=\sffamily\fontsize{7.4}{8.5}\selectfont}]
\path[use as bounding box] (0,0) rectangle (84,65.5);
\fill[lightblue,rounded corners=1mm] (0,0) rectangle (40.5,20.5);
\fill[lightorange,rounded corners=1mm] (43.5,0) rectangle (84,20.5);
\node[heading,text=blueaccent,anchor=north] at (20.25,2) {Theory knowledge};
\node[heading,text=orangeaccent,anchor=north] at (63.75,2) {Case reasoning};
\node[small,anchor=north,align=center,text width=35mm] at (20.25,7) {Concepts and rule relations\\14 categories};
\node[small,anchor=north,align=center,text width=35mm] at (63.75,7) {Chart-conditioned interpretation\\11 categories};
\IfFileExists{figures/generated/dataset-glance-art.png}{%
  \node[anchor=center] at (42,16.1) {\includegraphics[width=43mm,height=6mm,keepaspectratio]{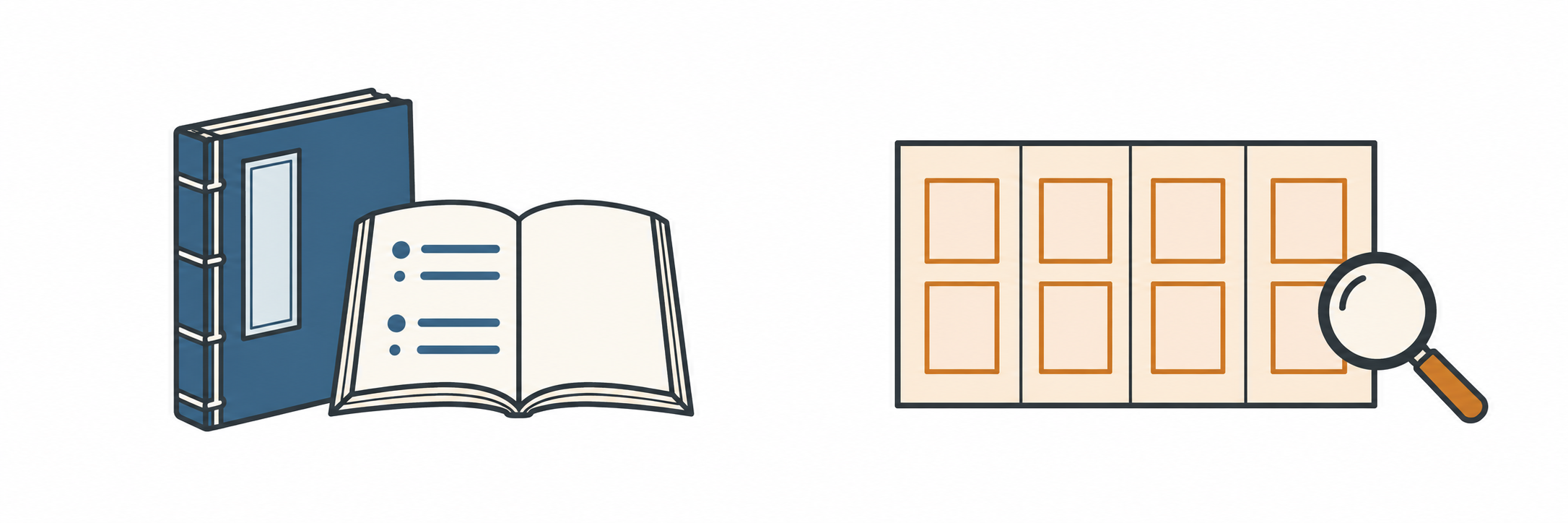}};
}{%
  \node[anchor=center,text=ink!65] at (42,16) {Conceptual artwork pending};
}
\node[heading,anchor=west] at (0,23) {Item distribution};
\node[anchor=west] at (0,29) {Original};
\node[anchor=west] at (0,37) {Final};
% Common count scale: 51 mm = 3,000 items; bars start at 19 mm.
\fill[blueaccent] (19,26) rectangle (44.5,32);
\fill[orangeaccent] (44.5,26) rectangle (70,32);
\node[text=white] at (31.75,29) {1,500};
\node[text=white] at (57.25,29) {1,500};
\fill[blueaccent] (19,34) rectangle ({19+51*(1454/3000)},40);
\fill[orangeaccent] ({19+51*(1454/3000)},34) rectangle ({19+51*(2492/3000)},40);
\node[text=white] at ({19+51*(1454/6000)},37) {1,454};
\node[text=white] at ({19+51*((1454+1038/2)/3000)},37) {1,038};
\node[anchor=east] at (84,29) {3,000};
\node[anchor=east,font=\sffamily\bfseries\fontsize{8}{9.3}\selectfont] at (84,37) {2,492};
\draw[blueaccent,line width=0.7pt] (0,44) -- (40.5,44);
\draw[orangeaccent,line width=0.7pt] (43.5,44) -- (84,44);
\node[anchor=north west,align=left,text width=39.5mm] at (0,46) {\textbf{Rule question}\par
  \small Explicit rule or relation\\[-1pt]
  \small Four options: A / B / C / D};
\node[anchor=north west,align=left,text width=40.5mm] at (43.5,46) {\textbf{Chart + interpretation}\par
  \small Four pillars + question\\[-1pt]
  \small Four options: A / B / C / D};
\end{tikzpicture}
\caption{Benchmark overview. Theory covers concepts and rule relations across 14 categories; Case covers chart-conditioned interpretation across 11 categories. Shared-scale bars show task composition in the original 3,000-item benchmark and final 2,492-item set. The miniature schematics illustrate the two question formats.}
\label{fig:dataset-glance}
\end{figure}

\section{Introduction}
% Second-page, full-width construction pipeline and representative examples.
% Chinese strings preserve the clean-release items (line breaks are typographic).
\begin{figure*}[t]
\centering
\begin{tikzpicture}[x=1mm,y=-1mm,
  every node/.style={font=\sffamily\fontsize{8}{9.3}\selectfont,text=ink,inner sep=0pt},
  flow/.style={draw=ink!25,fill=white,rounded corners=1mm,align=center,
    minimum height=10mm,text width=25mm,inner sep=1mm},
  title/.style={font=\sffamily\bfseries\fontsize{9}{10.5}\selectfont},
  arrow/.style={-{Stealth[length=1.4mm,width=1.0mm]},draw=ink!60,line width=0.55pt}]
\path[use as bounding box] (0,0) rectangle (174,80);
\IfFileExists{figures/generated/dataset-examples-art.png}{%
\node[anchor=center] at (10,7) {\includegraphics[width=14mm,height=6mm,keepaspectratio]{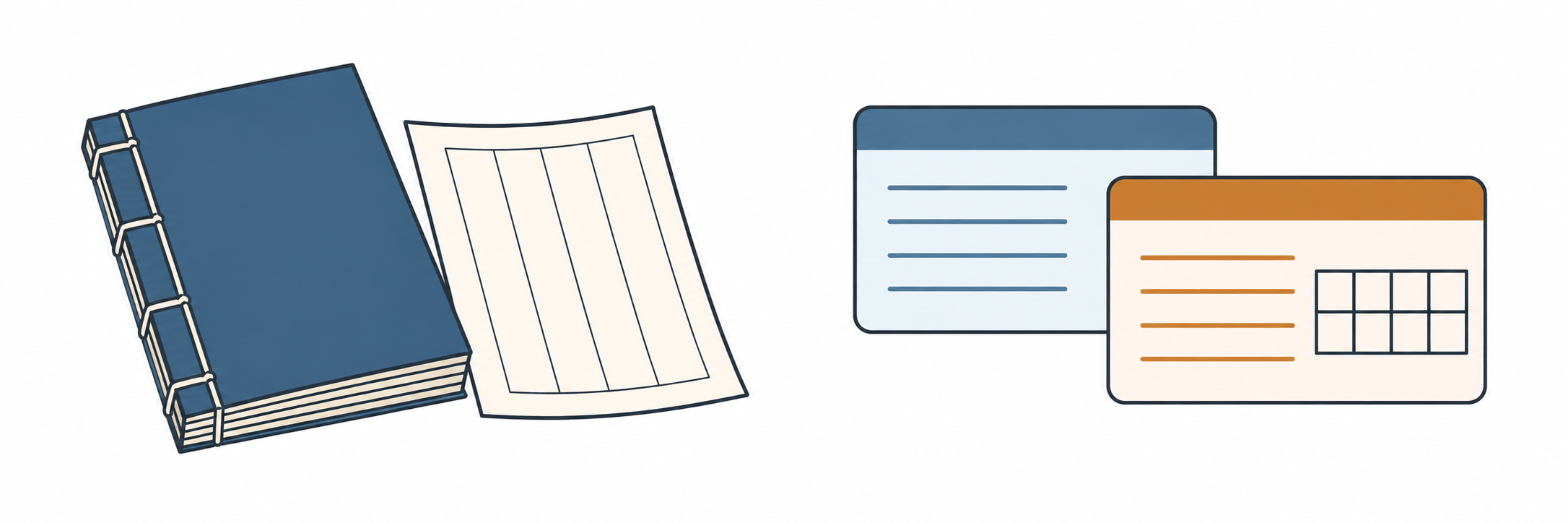}};
}{%
  \node[anchor=center,align=center,text=ink!65,text width=18mm] at (10,7) {Artwork\par pending};
}
\node[flow,text width=27mm] (s1) at (31,7) {Classical + instructional\par sources\par + extraction};
\node[flow] (s2) at (66,7) {Question creation\par + checking};
\node[flow] (s3) at (101,7) {Original benchmark\par \textbf{3,000 items}};
\node[flow,text width=27mm] (s4) at (136,7) {Model evaluation\par + model-informed\par refinement};
\node[flow,text width=24mm] (s5) at (165,7) {Final benchmark\par \textbf{2,492 items}};
\draw[arrow] (s1.east) -- (s2.west);
\draw[arrow] (s2.east) -- (s3.west);
\draw[arrow] (s3.east) -- (s4.west);
\draw[arrow] (s4.east) -- (s5.west);
\node[anchor=west,text=ink!75] at (0,16) {Chinese originals \enspace\textbullet\enspace English translations by the authors};
\node[anchor=west,text=ink!75,align=left,text width=174mm] at (0,19) {Refinement uses evaluated-model performance; the final set is not an independent holdout.};
% Examples use typography and rules rather than large colored cards.
\node[title,text=blueaccent,anchor=north west] at (3,24) {Theory example \quad {\cjkfont 五行}};
\node[title,text=orangeaccent,anchor=north west] at (92,24) {Case example \quad {\cjkfont 事业}};
\node[anchor=north west] at (3,29) {\texttt{mq4\_000001}};
\node[anchor=north west] at (92,29) {\texttt{me4\_000001}};
\draw[blueaccent!35] (3,33) -- (82,33);
\draw[orangeaccent!35] (92,33) -- (171,33);
\node[anchor=north west,align=left,text width=79mm] at (3,35) {
  {\cjkfont 午未合导致未土增力，这体现了五行中的哪种关系？}\par\smallskip
  The Wu--Wei combination strengthens Wei Earth. Which\par relationship among the Five Elements does this illustrate?\par\medskip
  \textbf{A}\ {\cjkfont 相克关系}\quad Mutual restraint\qquad
  \textbf{B}\ {\cjkfont 相刑关系}\quad Mutual punishment\par\smallskip
  \textbf{C}\ {\cjkfont 相生关系}\quad Mutual generation\qquad
  \textbf{D}\ {\cjkfont 相冲关系}\quad Mutual clash
};
\node[anchor=north west,align=left,text width=79mm] at (92,35) {
  {\cjkfont 【命例信息】性别：男}\quad Case information: male\par
  {\cjkfont 年柱丁巳\quad 月柱壬寅\quad 日柱辛酉\quad 时柱乙未}\par
  Year Ding-Si \quad Month Ren-Yin \quad Day Xin-You \quad Hour Yi-Wei\par\smallskip
  {\cjkfont 【问题】根据原命局，此命的事业身份最可能是什么？}\par
  Given the natal chart, what is this person's most likely career status?\par\smallskip
  \textbf{A}\ {\cjkfont 生意人}\ Businessperson\quad
  \textbf{B}\ {\cjkfont 做官的}\ Official\quad
  \textbf{C}\ {\cjkfont 艺术家}\ Artist\quad
  \textbf{D}\ {\cjkfont 普通的平民百姓}\ Ordinary citizen
};
\node[anchor=south west,text=ink!75,font=\sffamily\fontsize{7.4}{8.5}\selectfont] at (3,78) {Dataset key: C};
\node[anchor=south west,text=ink!75,font=\sffamily\fontsize{7.4}{8.5}\selectfont] at (92,78) {Dataset key: D};
\end{tikzpicture}
\caption{Construction pipeline and representative examples. Classical and instructional texts support extraction, question creation, and model-based checking. Evaluation of the original 3,000 items informs refinement to 2,492 retained items. The examples show Chinese questions with author English translations; keys are dataset labels. Refinement uses evaluated-model performance, so the final set is not an independent holdout.}
\label{fig:pipeline}
\end{figure*}

Can a language model that knows a domain's vocabulary and rules use them reliably when several rules meet in a case? This distinction matters for systems that retrieve, explain, or apply specialized cultural knowledge. A strong overall score can conceal a weaker ability to answer contextual questions, while an undifferentiated reasoning label can hide tasks that remain comparatively easy.

Traditional Chinese Bazi, or Four Pillars, provides a structured setting for evaluating culturally specific knowledge. We use it as a cultural knowledge representation, not as an empirically validated forecasting system. Our benchmark pairs \emph{Theory} questions about concepts and rule relations with \emph{Case} questions that supply chart context and request an interpretation. Figure~\ref{fig:dataset-glance} summarizes the task families, category coverage, and original and final item distributions. This organization supports task-specific comparisons of rule knowledge and contextual answers, while the two question lists do not isolate cognitive mechanisms.

Our central finding is a selective, rather than uniform, application gap. Five of six systems exceed 90\% on Twelve Stages and Nayin, yet all six remain below 45\% on Career and Family Relations cases. Conversely, Luck Pillars cases score 78.21--89.74\%. Thus neither uniform Bazi competence nor a universal inability to solve cases describes the observed profile. The practical question is which parts of a domain an aggregate score represents.

We make three contributions:
\begin{enumerate}
\item A source-derived benchmark spanning 25 categories, with separate Theory and Case tasks, an original 3,000-item release, and a documented 2,492-item evaluation subset.
\item A six-system analysis of task and category strengths, Case question subtypes, and invalid-response sensitivity, exposing limitations hidden by overall rankings.
\item Paired DeepSeek native/disabled configuration comparisons on the original set, showing that Theory gains do not imply comparable Case gains.
\end{enumerate}

The answer key is model-generated and model-verified, and refinement used evaluated outcomes. These boundaries remain explicit: our findings describe performance on this selected textual artifact, not verified expertise, a model-independent holdout, or the truth of Bazi predictions. Within that scope, the category profiles provide a more informative basis for evaluating domain applications than a single leaderboard score.

\section{Related Work}

\paragraph{Knowledge and cultural evaluation.}
MMLU, BIG-bench, TruthfulQA, and GPQA assess broad knowledge and challenging QA \citep{hendrycks2021mmlu,srivastava2023bigbench,lin2022truthfulqa,rein2023gpqa}; CLUTRR targets relational inference \citep{sinha2019clutrr}. CLUE, C-Eval, CMMLU, and GlobalBench broaden linguistic and cultural coverage \citep{xu2020clue,huang2023ceval,li2024cmmlu,song2023globalbench}. The adjacent BaZi-Based Character Simulation Benchmark studies temporal and persona reasoning \citep{zheng2025bazi}. Our contribution is a Theory/Case and category-level analysis within a four-choice task, not a first-Bazi-benchmark claim.

\paragraph{Evaluation boundaries.}
Model-in-the-loop benchmarks and contrast sets motivate explicit selection histories \citep{kiela2021dynabench,nie2020anli,gardner2020contrastsets}; targeting and contamination studies motivate bounded score interpretation \citep{schlangen2021targeting,sainz2023contamination,jacovi2023stopuploading,magar2022contamination}. Our item diagnostics are panel conditional, not an item-response-theory calibration \citep{lalor2016irt,lalor2018difficulty,vania2021irt}. Chain-of-thought, self-consistency, test-time compute, and process-supervision research distinguish inference configuration from answer correctness \citep{wei2022chain,wang2023selfconsistency,snell2024testtime,lightman2023process,suzgun2023challenging,zhou2026overthinking}. Accordingly, we report task-dependent configuration associations without claiming an isolated reasoning mechanism.

\section{Benchmark and Tasks}
\label{sec:construction}

\paragraph{What the benchmark asks.}
Each record has an identifier, a Chinese question or chart context, four options, an answer key, and task metadata. Category names organize content; they are not validated cognitive-skill labels. Figure~\ref{fig:pipeline} traces the construction process from source extraction and question checking to model evaluation and refinement. Its representative items make the task distinction concrete: the Theory item asks about a named rule relation, whereas the Case item supplies a four-pillar chart before requesting an interpretation. Both use four-option answers. The examples retain their Chinese content and dataset keys, with English translations provided by the authors.

\paragraph{Source-linked construction.}
Classical and instructional texts are converted to passages and knowledge units, from which questions are generated and checked for schema and answer consistency. DeepSeek chat participates in extraction, drafting, verification, and deduplication. Theory draws on 29 source works; 975/1,038 final Case items derive from four modern or instructional works. The public schema omits item-level source and school fields, and comprehensive expert adjudication is not documented. We therefore describe the key as model-generated and model-verified. Appendix~\ref{app:taxonomy} supplies the full taxonomy and provenance limits.

\paragraph{Model-informed refinement.}
After the original 3,000-item evaluation, item difficulty and discrimination informed category quotas and deletion priorities. The final set retains 2,492 items: 1,454 Theory and 1,038 Case. It removes 508 items (16.93\%), comprising 477 answered correctly by all six native systems, 30 by five systems, and one by none. Four overlapping conflict/near-duplicate reasons do not add to that count. Retained identifiers, questions, options, and keys are unchanged, so final scores filter archived responses rather than rerun models. The procedure is explicitly \emph{model-informed}: it does not create an independent holdout. Final scores and intervals describe a selected list, with no demonstrated post-selection error guarantees. We retain original-set comparisons as the pre-selection reference. The change also increases the Theory share from 50\% to 58.35\%, motivating task and category breakdowns alongside overall scores. The original release, fixed-seed builder, and deletion manifest preserve the selection history. Appendix~\ref{app:hardening} gives the rules, before/after distributions, and panel-conditional discrimination changes.

\section{Evaluation Setup}
\label{sec:setup}

\paragraph{Systems and scoring.}
We evaluate Kimi-K3, GLM-5.3, Qwen3.8-Max, MiniMax-M3, DeepSeek-V4.1-Flash, and DeepSeek-V4-Pro-0813. We call them \emph{endpoint systems}: four use mixed API routes and immutable model snapshots are unavailable. Each item receives a zero-shot request at temperature zero, with no tools, retrieval, demonstrations, or explicit chain-of-thought request. The visible prompt requests one answer letter. Accuracy counts invalid or empty canonical answers as wrong; the four-choice random baseline is 25\%.

\paragraph{Comparisons.}
Main results use the final 2,492 items. Category means average six endpoint accuracies, computed from correct counts and denominators; category ranks are descriptive, not significance tests. Nominal Wilson intervals and paired McNemar tests describe the observed lists. Holm correction covers 15 original-set model pairs. Source/chart clustering remains unmodeled, and final-set inference additionally conditions on model-informed selection.

DeepSeek configuration comparisons instead use the original 3,000 items. Native means no thinking-disable parameter was sent; the counterpart sends \code{thinking: \{"type":"disabled"\}} with the same visible prompt. Six Overall/Theory/Case tests form a separate Holm family. Appendix~\ref{app:prompt} records parsing, retries, routes, token limits, and statistical definitions. No new evaluation runs were performed for this analysis.

\section{Results and Analysis}
\label{sec:main-results}

\subsection{Overall scores conceal a task gap}

Table~\ref{tab:leaderboard} reports final-set accuracy. Scores range from 63.20\% to 75.56\%, with Kimi-K3, GLM-5.3, and Qwen3.8-Max forming the highest-scoring group. These are operational endpoint results under invalid-as-wrong scoring, not provider-neutral capability rankings. On the original 3,000 items, Holm-corrected paired tests detect differences in 11/15 pairs but not among the top three or between the DeepSeek aliases; non-rejection does not establish equivalence. Full pairwise and original-set results appear in Appendix~\ref{app:robustness}.

\begin{table}[t]
\centering
\caption{Post-selection results on the final benchmark candidate (\%). Accuracy includes invalid answers; intervals are nominal Wilson 95\% descriptions.}
\label{tab:leaderboard}
\footnotesize
\setlength{\tabcolsep}{4pt}
\begin{tabular}{lrrrrr}
\toprule
System & Overall & 95\% CI & Theory & Case & Invalid \\
\midrule
Kimi-K3 & 75.56 & 73.84--77.21 & 86.04 & 60.89 & 4.98 \\
GLM-5.3 & 75.08 & 73.34--76.74 & 84.59 & 61.75 & 0.24 \\
Qwen3.8-Max & 74.84 & 73.10--76.50 & 84.59 & 61.18 & 5.98 \\
DS V4.1 Flash & 72.43 & 70.64--74.15 & 83.56 & 56.84 & 0.00 \\
DS V4 Pro & 72.35 & 70.56--74.07 & 84.66 & 55.11 & 0.00 \\
MiniMax-M3 & 63.20 & 61.29--65.07 & 71.25 & 51.93 & 1.08 \\
\bottomrule
\end{tabular}
\end{table}

\begin{figure}[t]
\centering
\tightfig{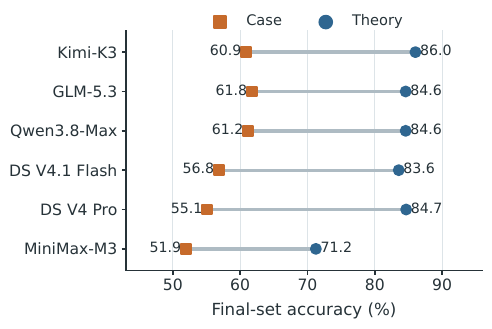}
\caption{Final-set Theory and Case accuracy. Every endpoint has a lower Case score; invalid-response sensitivity retains the gap.}
\label{fig:theory-case}
\end{figure}

Every system scores higher on Theory than Case. Excluding MiniMax, Theory spans 83.56--86.04\%, while Case spans 55.11--61.75\%; MiniMax scores 71.25\% and 51.93\%, respectively. Figure~\ref{fig:theory-case} shows that the same aggregate direction holds across all six endpoints, despite different absolute levels.

Completion failures amplify, but do not account for, this contrast. Of 306 final-set invalid responses, 258 occur on Case items. Excluding invalids separately within each task leaves gaps of 16.60--29.56 points. Kimi and Qwen then reach 68.77\% and 69.47\% on Case, still below their 86.34\% and 86.07\% Theory scores. This is a sensitivity analysis, not a replacement leaderboard: it conditions on successful completion.

Task composition also matters. Refinement increases the weight of the higher-scoring Theory task. Equal-weight task-macro scores are therefore 1.61--2.47 points below micro scores; category-macro scores are 1.38--2.20 points lower, with unchanged ordering. An application-oriented evaluation should report its task mix alongside any aggregate score.

\subsection{Theory: uneven command of domain rules}

\begin{figure*}[t]
\centering
\tightfig[\textwidth]{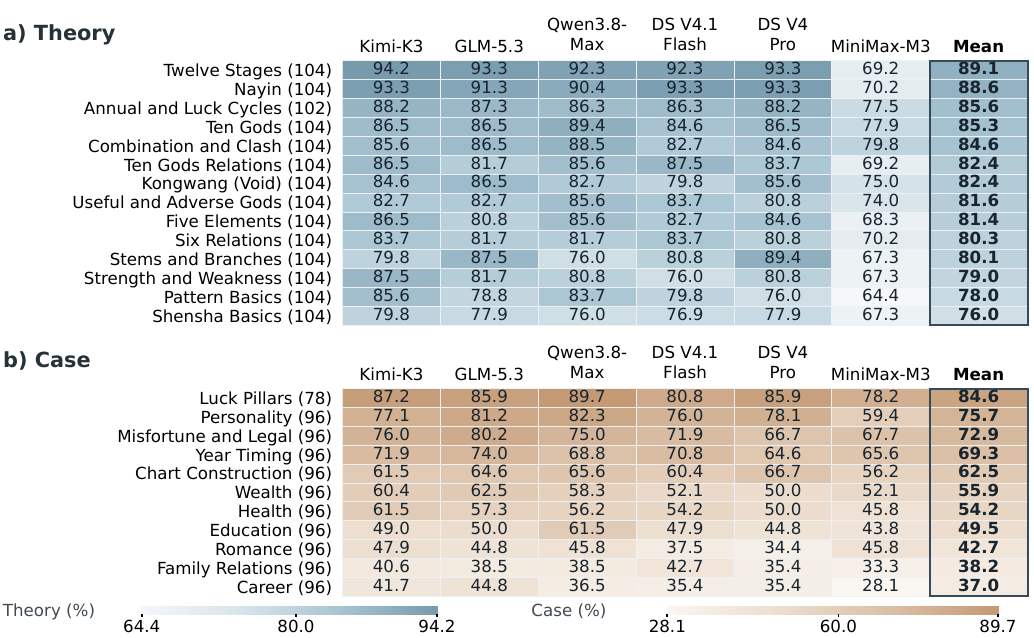}
\caption{Final-set category accuracy (\%), shown to one decimal place. The outlined Mean column reports the six-model mean computed from unrounded accuracies. Categories are sorted by six-model mean; numbers in parentheses indicate the number of items evaluated per model. These are descriptive post-selection results.}
\label{fig:category-profile}
\end{figure*}

Figure~\ref{fig:category-profile} shows endpoint accuracies for all 25 categories in an annotated heatmap; the final column summarizes the six-model mean. Cells display one decimal place; the text retains two. Twelve Stages and Nayin are the highest-scoring Theory categories, averaging 89.10\% and 88.62\%. Five systems exceed 90\% on both; MiniMax scores 69.23\% and 70.19\%. Thus the panel-level strength is substantial but not universal. Annual and Luck Cycles also scores comparatively well at 85.62\%.

The lower-scoring Theory categories are Shensha Basics (75.96\%), Pattern Basics (78.04\%), and Strength and Weakness (79.01\%). The 13.14-point difference between Twelve Stages and Shensha Basics shows why high Theory accuracy should not be read as uniformly reliable rule knowledge. Even within the stronger task family, category-level checking remains necessary.

Model differences further complicate an overall ranking. DeepSeek Pro reaches 89.42\% on Stems and Branches, compared with Qwen's 75.96\%, whereas Qwen leads the panel on Ten Gods at 89.42\%. Kimi reaches 87.50\% on Strength and Weakness and 85.58\% on Pattern Basics. These are observed category point estimates, not evidence of statistically established specialization or a stable model-selection rule.

Nor do Theory labels cleanly divide recall from reasoning. The existing question-format tag \emph{reasoning} yields scores of 75.35--91.67\%, while \emph{judgement} yields 66.90--83.62\%. A label alone does not establish reasoning depth. Differences can reflect item content, source conventions, answer-key construction, or format; this benchmark does not isolate those mechanisms.

\subsection{Case: selective rather than universal difficulty}

Case performance varies more sharply. Luck Pillars averages 84.62\%, followed by Personality at 75.69\% and Misfortune and Legal at 72.92\%. In contrast, Career averages 36.98\%, Family Relations 38.19\%, and Romance 42.71\%. All six systems remain below 45\% on Career and Family Relations, whereas all exceed 78\% on Luck Pillars. The range between the highest- and lowest-scoring Case categories is 47.64 points.

This pattern rules out a simple summary that contextual questions are uniformly difficult. It also requires care in naming the strength: the Luck Pillars category includes interpretations of luck periods, not only mechanical chart calculations. High scores there do not demonstrate a separately measured calendrical algorithm, just as low Career scores do not identify a particular failed reasoning step.

There is no single category winner. Qwen reaches 61.46\% on Education, compared with 43.75--50.00\% for the other systems, while Flash has the highest Family Relations point estimate at 42.71\%. That latter distinction is modest in practical terms because the entire panel performs poorly on the category. The useful application-facing conclusion is the profile of limits, not the name of a winner.

\subsection{Case subtypes and shared failures}

The Case metadata distinguishes questions tagged \emph{fact} from those tagged \emph{theory}. We describe them as \emph{recorded-outcome} and \emph{rule-interpretation} questions: the former refer to events represented in the source case, whereas the latter ask for a domain-rule interpretation. These are dataset labels, not independently verified facts or validated measures of two cognitive abilities.

% Generated by scripts/build-category-analysis.py; do not edit values.
\begin{table}[t]
\centering
\caption{Accuracy by existing Case label (\%). ``Fact'' and ``theory'' are dataset tags, not verified outcomes. Gap = theory minus fact, using unrounded values.}
\label{tab:case-subgroups}
\footnotesize
\setlength{\tabcolsep}{4pt}
\begin{tabular}{lrrr}
\toprule
System & \shortstack{Fact\\($n=464$)} & \shortstack{Theory\\($n=574$)} & Gap (pp) \\
\midrule
Kimi-K3 & 55.60 & 65.16 & +9.55 \\
GLM-5.3 & 56.03 & 66.38 & +10.34 \\
Qwen3.8-Max & 56.47 & 64.98 & +8.52 \\
DS V4.1 Flash & 51.29 & 61.32 & +10.03 \\
DS V4 Pro & 45.91 & 62.54 & +16.64 \\
MiniMax-M3 & 45.04 & 57.49 & +12.45 \\
\bottomrule
\end{tabular}
\end{table}

All six systems score lower on the 464 recorded-outcome items than on the 574 rule-interpretation items (Table~\ref{tab:case-subgroups}). The gaps, computed before rounding, range from 8.52 to 16.64 points. Even a model that handles source-domain rules relatively well may not recover a case's reported outcome. However, the two subsets differ in content and category composition; the association is not a controlled test of whether outcome questions themselves cause errors.

Agreement statistics reveal another limitation of averages. The final set contains 1,118 items answered correctly by all six systems and 204 answered correctly by none: 44.86\% and 8.19\% of the release. Case has fewer unanimous-correct items (25.3\%) and more unanimous-wrong items (13.2\%) than Theory (58.8\% and 4.6\%). Career and Family Relations each have no unanimous-correct items and 21.88\% unanimous-wrong items. Agreement with a shared wrong answer is not required by this statistic; it counts failure to match the key.

These are candidate strata for expert review, not a causal error taxonomy. Without adjudicated answers and coded explanations, we cannot distinguish missing knowledge, competing school conventions, ambiguous questions, or flawed keys from answer vectors alone. For a domain application, separate validation of rule explanation, contextual interpretation, and output completion is more informative than assuming that one high score transfers to all three.

\section{Reasoning Configuration Analysis}
\label{sec:ablation}

\subsection{Theory gains do not imply Case gains}

Table~\ref{tab:ablation} compares DeepSeek native and disabled configurations on the original 3,000 items, before selection using native outcomes. Flash's overall difference is +1.43 points and is not detected after Holm correction; Pro's is +6.73 points. The aggregate changes conceal different task associations: native Flash gains 6.53 points on Theory but loses 3.67 on Case, whereas native Pro gains 12.20 on Theory and only 1.27 on Case. Pro's Case difference is not detected after correction.

\begin{table}[t]
\centering
\caption{DeepSeek native versus disabled provider-configuration contrast on the original 3,000 items. Delta is native minus disabled accuracy in percentage points; Holm correction covers the six displayed tests. Native Flash records used 32,768 and later 65,536 maximum output tokens; native Pro used 65,536 throughout.}
\label{tab:ablation}
\footnotesize
\setlength{\tabcolsep}{2.3pt}
\begin{tabular*}{\columnwidth}{@{\extracolsep{\fill}}llrrr>{\raggedleft\arraybackslash}p{1.62cm}@{}}
\toprule
System & Subset & Native & Disabled & $\Delta$ & Holm $p$ \\
\midrule
Flash & Overall & 76.97 & 75.53 & +1.43 & $0.1795$ \\
Flash & Theory & 84.00 & 77.47 & +6.53 & $1.66 \times 10^{-8}$ \\
Flash & Case & 69.93 & 73.60 & -3.67 & $0.00935$ \\
Pro & Overall & 76.83 & 70.10 & +6.73 & $3.26 \times 10^{-13}$ \\
Pro & Theory & 85.07 & 72.87 & +12.20 & $4.92 \times 10^{-24}$ \\
Pro & Case & 68.60 & 67.33 & +1.27 & $0.3663$ \\
\bottomrule
\end{tabular*}
\end{table}

Both Theory contrasts are positive and detected, yet neither system shows a correspondingly detected positive Case difference. Additional native reasoning behavior is therefore not a uniform remedy for the observed application gap. This result supports reporting task-specific configuration comparisons instead of interpreting a total-score gain as improvement across a domain.

The paired records also change answers in both directions. Flash matches the key only in the native configuration on 328 items but only in the disabled configuration on 285; Pro has 465 versus 263. These transitions explain the net gains without identifying why a particular answer changed. No individual failure mechanism is inferred from them.

\subsection{What the configuration contrast can establish}

The visible prompt is shared, but the requests are not an isolated intervention on a reasoning mechanism. Native input counts contain fixed offsets of 26 tokens per Flash item and 79 per Pro item. Their contents are not exposed in the canonical table. Native Flash's maximum output limit changed from 32,768 to 65,536 during execution, and immutable provider snapshots are absent. We therefore report provider-configuration associations, not causal reasoning effects. Comparing native Flash with native Pro would instead be a cross-system comparison.

Operational costs are also asymmetric. On the original set, median native/disabled request latency is 10.68/1.43 seconds for Flash and 20.95/2.07 seconds for Pro. Those durations include network and queueing as well as provider execution; configurations ran at different times. They are not compute-only estimates or monetary costs. Appendix~\ref{sec:efficiency} retains the token, latency, and route evidence.

For application design, these observations motivate a bounded evaluation strategy: compare configurations on the actual task categories and check whether additional latency accompanies useful answer improvements. This study does not establish an optimal configuration policy. It shows why assuming that a reasoning-enabled configuration helps every kind of cultural-domain question is not supported by the observed results.

\section{Limitations}
\label{sec:limitations}

\paragraph{Representation and validation.}
The model-generated, model-verified key lacks comprehensive expert adjudication. Contested school conventions, incomplete public item provenance, concentrated Case sources, and unknown contamination limit interpretation. Scores measure agreement with this representation, not metaphysical truth, predictive validity, or general cultural competence. Applications involving consequential real-world decisions are outside the supported scope.

\paragraph{Selection and statistical scope.}
Final items were selected using the evaluated panel, not held out independently. Categories and Case tags are observational groupings; source and chart reuse remain unmodeled. Point estimates and nominal intervals do not establish population-level category effects. Shared generation/evaluation model families and single runs at temperature zero add uncertainty.

\paragraph{Endpoint limitations.}
Mixed routes, missing immutable snapshots, and configuration offsets restrict capability and efficiency comparisons.

\section{Conclusion}

Bazi evaluation reveals a more specific picture than a single knowledge or reasoning score. Models score well on some Theory categories but unevenly on others; Case performance ranges from strong Luck Pillars scores to shared weaknesses in Career and Family Relations. Paired DeepSeek configuration changes improve Theory more consistently than Case. These findings favor task-specific validation for cultural-domain applications. The benchmark supplies a documented setting for that analysis, while model-informed selection and answer-key limitations keep the conclusions conditional on the evaluated artifact.

\section*{Disclosures}
\paragraph{Resources and data availability.}
BaZi2500 is publicly available on \href{https://huggingface.co/datasets/MonsterPPPPP/BaZi2500}{Hugging Face} under CC BY-NC 4.0. Evaluation code and reproducibility resources are available in the \href{https://github.com/MonsterPPPP/bazi-qa-benchmark}{GitHub repository}; benchmark results and documentation are provided on the \href{https://monsterpppp.github.io/bazi-qa-benchmark/}{project page}. The authors have confirmed source redistribution rights and completed privacy review. This textual study recruited no participants. Scores must not support consequential real-world decisions (Appendix~\ref{app:reproducibility}).
\paragraph{Funding and competing interests.}
This work was supported by Beijing Liuyi Guanhua Technology Co., Ltd. The authors declare no competing interests.
\paragraph{AI use.}
Models assisted benchmark construction and preparation of this manuscript. They did not provide independent expert validation. Numerical results derive from checked analysis artifacts.

\section*{Implications for Application}
The results suggest a practical evaluation rule for systems that work with
culturally specific knowledge: test retrieval of named rules separately from
their use in a structured situation. A system can be highly reliable on a
definition or relation while failing when several chart fields must be combined
before selecting an answer. In this benchmark, Twelve Stages and Nayin are
the clearest Theory strengths, whereas Shensha Basics, Pattern Basics, and
Strength and Weakness are less uniform. The Case profile is sharper still:
Luck Pillars is comparatively strong, but Career, Family Relations, and
Romance remain difficult across the evaluated systems. These are descriptive
diagnostics for the released answer key, not evidence that any category is a
valid real-world predictor.

For an application designer, a single aggregate score should therefore be
treated as a routing signal rather than a readiness certificate. A useful
evaluation dashboard should expose task family, category, invalid-response
rate, and whether the prompt requests a factual relation or an interpretation.
The native/disabled comparisons reinforce this point: additional provider-side
reasoning behavior was associated with larger Theory gains than Case gains, but
the configurations also differed in token limits, routes, and execution time.
The benchmark can reveal that mismatch and motivate targeted tests; it cannot
identify a universal reasoning mechanism or prescribe a production setting.

Finally, the final list is a model-informed refinement of the original 3,000
items. It is useful for describing the retained panel, but it is not an
independent holdout. Applications should report which release, prompt, answer
key, and endpoint configuration they use, preserve invalid outputs for audit,
and avoid treating a correct multiple-choice label as expert certification.

\paragraph{What the benchmark can diagnose.}
The category profile is most useful as a prioritization device. High scores on
Twelve Stages or Nayin indicate that the model can often map a named concept to
one of four released labels. They do not show that it can reconcile competing
relations, identify which chart fields are relevant, or explain an answer in a
new context. Conversely, low scores on Career or Family Relations identify a
stress test for chart-conditioned interpretation, but they do not by themselves
separate missing domain knowledge from ambiguity in the key. Those hypotheses
require item-level adjudication and controlled prompt variants, which are
outside this release.

\paragraph{Responsible use.}
The benchmark is appropriate for comparing text systems, monitoring regressions,
and selecting cases for expert review. It is not a certification test for
practitioners, a basis for advice about a person, or evidence that a model has
learned a metaphysical system in a historically complete sense. The answer key
is a property of this release. Any downstream application should expose that
status to users, retain the underlying question and answer trace, and route
high-stakes or contested interpretations to qualified human review. These
practices turn the observed Theory--Case mismatch into an engineering signal
without converting a descriptive benchmark result into a claim about people or
the world.

\paragraph{Reading a benchmark score.}
A score measures released-key agreement, not metaphysical truth, forecasting
validity, or expert certification. Theory emphasizes rule knowledge; Case
adds chart-conditioned interpretation, but these are not paired tests isolating
cognitive mechanisms. Invalid responses count as wrong; excluding them
conditions accuracy on completion. The 2,492 retained items were selected from
the original 3,000 using model outcomes. Final-set scores are therefore
post-selection descriptions; the original list remains the pre-selection
reference.

\paragraph{Minimum reporting record.}
Bind Theory/Case scores and category denominators to the release, key version,
item IDs, prompt, parser, retry rule, and endpoint configuration. Include invalid
and empty-response rates, token limits, run dates, and response traces showing
how retries become scored records. Configuration contrasts also need a stated
difference direction and correction family. This record makes scores auditable;
an aggregate alone cannot replace task-specific diagnostics.

\label{body:end}
\clearpage
\label{refs:start}
\bibliographystyle{plainnat}
\begingroup
\raggedright
\bibliography{references}
\endgroup
\label{refs:end}
\clearpage
\appendix
\section{Dataset and Provenance}
\label{app:start}
\label{app:taxonomy}

\begin{table}[t]
\centering
\caption{Dataset statistics for the original and final releases.}
\label{tab:dataset-stats}
\small
\begin{tabular}{lrr}
\toprule
Statistic & Original & Final \\
\midrule
Total items & 3,000 & 2,492 \\
Theory items & 1,500 & 1,454 \\
Case items & 1,500 & 1,038 \\
Theory categories & 14 & 14 \\
Case categories & 11 & 11 \\
Options per item & 4 & 4 \\
Removed items & -- & 508 (16.93\%) \\
\bottomrule
\end{tabular}
\end{table}

\begin{table*}[t]
\centering
\caption{Final taxonomy and category sizes.}
\label{tab:taxonomy}
\scriptsize
\begin{tabularx}{\textwidth}{lYrlYr}
\toprule
ID & Theory category & $n$ & ID & Case category & $n$ \\
\midrule
T01 & Five Elements & 104 & C01 & Career & 96 \\
T02 & Stems and Branches & 104 & C02 & Education & 96 \\
T03 & Ten Gods & 104 & C03 & Wealth & 96 \\
T04 & Ten Gods Relations & 104 & C04 & Romance & 96 \\
T05 & Combination and Clash & 104 & C05 & Health & 96 \\
T06 & Strength and Weakness & 104 & C06 & Personality & 96 \\
T07 & Pattern Basics & 104 & C07 & Year Timing & 96 \\
T08 & Useful and Adverse Gods & 104 & C08 & Chart Construction & 96 \\
T09 & Shensha Basics & 104 & C09 & Luck Pillars & 78 \\
T10 & Nayin & 104 & C10 & Misfortune and Legal & 96 \\
T11 & Kongwang (Void) & 104 & C11 & Family Relations & 96 \\
T12 & Twelve Stages & 104 & & & \\
T13 & Annual and Luck Cycles & 102 & & & \\
T14 & Six Relations & 104 & & & \\
\bottomrule
\end{tabularx}
\end{table*}

Table~\ref{tab:dataset-stats} summarizes the original 3,000-item benchmark and its final 2,492-item subset. Table~\ref{tab:taxonomy} gives the final English category names used in the figures. Theory's ``Six Relations'' tests conceptual kinship relations, whereas Case's ``Family Relations'' asks for case interpretation. Category-level analyses are descriptive unless otherwise marked; no family-wise hypothesis test is claimed for the heatmaps.

\subsection{Construction audit}

The original benchmark evaluated here contains 3,000 multiple-choice items: 1,500 Theory questions and 1,500 Case questions. Its construction process converts books into chapter text, extracts fine-grained source passages, drafts questions with provenance, and checks schema and answer uniqueness. The pipeline requires source evidence for an admitted item and records links between an item and its source paragraph or knowledge unit. A language model (DeepSeek chat) participates in gating, extraction, drafting, verification, and deduplication. The evidence audit found no comprehensive expert-human validation pass for all gold answers. We therefore use ``model-generated and model-verified'' rather than ``expert-annotated'' as the operative description.

The present paper audits and evaluates this existing 3,000-item benchmark and its deletion-only refinement. It does not claim that the paper package alone can regenerate the historical artifact from raw books. Historical prompts and registry versions are not complete in the available evidence.

The source mix is heterogeneous. The Theory portion spans 29 source works, including classical texts. In the final Case portion, 975 of 1,038 items (93.93\%) come from four modern or instructional works. The public JSONL schema strips per-item source and school fields, so a user of the release can reproduce the task and metadata but cannot reconstruct the full item-to-source map from the release alone. We therefore describe the resource as source-linked during construction and source-derived with incomplete public provenance, rather than claiming that every item is directly grounded in a classical passage.

\subsection{Category coverage and schema}

The taxonomy contains 14 Theory categories and 11 Case categories. Theory categories cover Five Elements, Stems and Branches, Ten Gods, relations, combinations and clashes, strength, patterns, useful and adverse gods, auxiliary signs, Nayin, void periods, Twelve Stages, annual/luck cycles, and family relations. Case categories cover career, education, wealth, romance, health, personality, year timing, chart construction, luck pillars, misfortune/legal matters, and family relations. The final release contains 104 items in each Theory category except Annual and Luck Cycles (102 after forced defect removal), and 96 items in ten Case categories; Luck Pillars contains all 78 available items.

No text is edited during release hardening. The original identifiers are retained, and the clean release is a strict subset of the original 3,000 records. This permits direct filtering of the original model responses rather than a second, potentially non-comparable model run.

The public release has near-balanced gold answer positions (A: 646, B: 607, C: 624, D: 615), and each native run has 2,492 unique identifiers, zero missing records, and zero cross-model metadata disagreements. These structural checks do not validate gold-answer correctness.

\section{Hardening Audit}
\label{app:hardening}
\label{sec:hardening}

\subsection{Selection rule and retained release}
Hardening follows full evaluation of the original 3,000 items. Difficulty is one minus the fraction of six native systems answering correctly; discrimination combines item-rest correlation and panel response spread (Appendix~\ref{app:statistics}). Both diagnostics are conditional on this small model panel.

After removing confirmed conflicts and near duplicates, the script targets 104 items per Theory category and 96 per Case category, retaining underfilled categories. Within an overfull category, deletion priority is six-of-six correct, five-of-six, one-of-six, two-to-four, then zero-of-six. Seed 20260922 breaks ties within bands. Near-duplicate retention also uses model-derived discrimination. The original release, fixed-seed script, and deletion manifest preserve this process.

The final release removes 508 unique items (16.93\%). The mutually exclusive removal-band distribution is 477 six-of-six correct, 30 five-of-six correct, and one zero-of-six correct. The same manifest includes four forced data-quality removals: two records with an identical stem but conflicting gold labels and two near-duplicate variants of a retained item. Those reason labels are embedded in the performance-band counts and must not be added as disjoint counts. Unanimous-correct items provide no discordant evidence for any model pair's McNemar test, motivating their priority. Because other bands are also deleted, we do not claim that every deletion is inferentially neutral.

Two category exceptions remain after quota balancing. Annual and Luck Cycles has 102 Theory items because two confirmed defective records were removed, and Luck Pillars has 78 Case items because the original category contained only 78. Every other Theory category has 104 items and every other Case category has 96.

\begin{table}[t]
\centering
\caption{Observed changes under model-informed hardening. Final-set test counts are post-selection descriptions.}
\label{tab:hardening}
\small
\begin{tabular}{lrr}
\toprule
Metric & Original & Final \\
\midrule
Items & 3,000 & 2,492 \\
Mean empirical difficulty & 0.2326 & 0.2776 \\
Mean discrimination & 0.2182 & 0.2574 \\
Six-model all-correct & 1,595 & 1,118 \\
Six-model all-wrong & 205 & 204 \\
Holm-significant pairs & 11/15 & 11/15 \\
\bottomrule
\end{tabular}
\end{table}

\begin{table}[t]
\centering
\caption{Actual deletion-band composition of the final hardening.}
\label{tab:removals}
\small
\begin{tabular}{lr}
\toprule
Item band & Removed \\
\midrule
Six of six correct & 477 \\
Five of six correct & 30 \\
Two to four correct & 0 \\
One of six correct & 0 \\
Zero of six correct & 1 \\
\midrule
Total & 508 \\
\bottomrule
\end{tabular}
\end{table}

\begin{table}[t]
\centering
\caption{Observed endpoint-system accuracy before and after hardening. Original scores are recomputed from the outcome matrix used by hardening; final scores filter those responses by retained ID.}
\label{tab:hardening-model-shift}
\scriptsize
\begin{tabular}{lrrr}
\toprule
Model & Original & Final & Change \\
\midrule
Kimi-K3 & 79.50 & 75.56 & -3.94 \\
GLM-5.3 & 79.13 & 75.08 & -4.05 \\
Qwen3.8-Max & 78.97 & 74.84 & -4.13 \\
DS V4.1 Flash & 76.97 & 72.43 & -4.54 \\
DS V4 Pro & 76.83 & 72.35 & -4.48 \\
MiniMax-M3 & 69.07 & 63.20 & -5.87 \\
\bottomrule
\end{tabular}
\end{table}

\subsection{Observed shifts and inference limits}
Table~\ref{tab:hardening} reports an increase in mean empirical difficulty from 0.2326 to 0.2776 and in mean discrimination from 0.2182 to 0.2574; the observed number of Holm-significant pairs remains 11/15. Filtering cached responses lowers all six accuracies by 3.94--5.87 percentage points (Table~\ref{tab:hardening-model-shift}). Category quotas reduce large Case categories, changing the task mix from 50/50 to 58.35\% Theory and 41.65\% Case. The removal summary in Table~\ref{tab:removals} uses mutually exclusive performance bands; overlapping data-quality reasons are described above.

These changes establish a harder selected list for this panel, not greater validity or universal difficulty. Outcome-informed inclusion can retain panel-specific weaknesses and distort subsequent comparisons. Final-set intervals and tests have no demonstrated post-selection coverage or error control; we use them descriptively and retain original-set comparisons as the pre-selection reference. An independent future evaluation requires inclusion decisions made before observing the evaluated systems.

The hardening report also includes exploratory schemes A, B, and C. Those schemes are simulations, not the final release. Separate threshold-sensitivity experiments showed that deleting five-of-six items could reduce detected pairs from 11/15 to 8/15 or 6/15, whereas the quota-based final subset retains 11/15 nominally detected pairs. Simulated sizes and scores are not used as final benchmark results.

\section{Evaluation Protocol and Statistical Definitions}
\label{app:prompt}

The main and disabled-configuration runs use the following user-facing template, with the question and four options substituted for each independent request:

\begin{minipage}{\columnwidth}
\begin{quote}\small\ttfamily\raggedright
You are answering a multiple-choice question about traditional Chinese Bazi theory.\\[2pt]
Select the single correct answer.\\
Output only one letter: A, B, C, or D.\\
Do not provide any explanation or additional text.\\[2pt]
Question:\\
\{question\}\\[2pt]
A. \{option\_A\}\\
B. \{option\_B\}\\
C. \{option\_C\}\\
D. \{option\_D\}
\end{quote}
\end{minipage}

\begin{table}[ht]
\centering
\caption{Run configuration. ``Native'' means no thinking-disable parameter was sent.}
\label{tab:run-config}
\scriptsize
\begin{tabular}{llll}
\toprule
Run family & Models & Reasoning & Source \\
\midrule
V1 native & Six listed models & Native & Mixed routes \\
V2 Flash & DS V4.1 Flash & Disabled & DeepSeek official \\
V2 Pro & DS V4 Pro & Disabled & DeepSeek official \\
\bottomrule
\end{tabular}
\end{table}

\begin{table*}[t]
\centering
\caption{Per-system run and route manifest. Raw counts include transport retries; canonical counts are the item records used by the analysis. Exact route segments, API model IDs, base-path classes, response-file paths, and complete hashes are retained in the recovered CSV manifest.}
\label{tab:run-route-manifest}
\scriptsize
\begin{tabularx}{\textwidth}{l X l l r l r l}
\toprule
Endpoint configuration & Observed route(s) & Mode/max & UTC span & Raw & Set & Canonical & SHA prefix \\
\midrule
DS V4 Pro (native) & DeepSeek official (3004) & native; 65536 & 13:32--16:21 & 3004 & final & 2492 & aa380761e51f \\
DS V4.1 Flash (native) & DeepSeek official (3002) & native; 32768/65536 & 09:07--10:02 & 3002 & final & 2492 & 2225eb6179df \\
MiniMax-M3 (native) & SCNet (1807); early/unknown (1340) & native; 32768/65536 & 07:45--13:24 & 3147 & final & 2492 & 981c54a982f4 \\
Kimi-K3 (native) & SCNet (1816); micuapi (1424); early/unknown (919) & native; 32768/65536 & 07:46--17:00 & 4159 & final & 2492 & c8b94ded91c9 \\
Qwen3.8-Max (native) & SCNet (1636); micuapi (1589); early/unknown (883) & native; 32768/65536 & 07:45--17:35 & 4108 & final & 2492 & 248c1ba7d807 \\
GLM-5.3 (native) & SCNet (1489); micuapi (1706); early/unknown (913) & native; 32768/65536 & 07:45--14:39 & 4108 & final & 2492 & ce25699f48ab \\
DS V4.1 Flash (disabled) & DeepSeek official (3000) & disabled; 65536 & 14:45--14:47 & 3000 & original & 3000 & 25d9a60a6a5e \\
DS V4 Pro (disabled) & DeepSeek official (3000) & disabled; 65536 & 14:35--14:37 & 3000 & original & 3000 & 27e44a4c73f1 \\
\bottomrule
\end{tabularx}
\end{table*}

All runs use temperature zero, no tools, no retrieval, no demonstrations, and one independent request per item. Original runs cover 3,000 items per system; the final table filters these to 2,492 identifiers. Disabled runs allow 65,536 output tokens and send this parameter:
\begin{quote}\small\ttfamily\raggedright
thinking: \{"type":"disabled"\}
\end{quote}
Native Flash began with a 32,768-token limit and later used 65,536; native Pro used 65,536. No canonical response has a \code{length} finish reason. Provider model names and API identifiers are retained when available, but not all systems have immutable snapshot identifiers.

The parser contract and full route/configuration manifest are paper-local files:
\begin{quote}\small\raggedright
\path{evidence/canonicalization_contract.md}\\
\path{tables/source/run_route_manifest.csv}
\end{quote}
The manifest records hashes and raw/canonical counts.

\subsection{Parsing, retries, and canonicalization}
The parser removes balanced reasoning tags and accepts a single A--D letter or an explicit answer marker. Empty, malformed, or multiple-answer outputs are invalid. Transport errors are excluded; among non-error records, the latest non-empty parsed answer is selected, otherwise the latest empty response. Raw retries remain in the manifest; accuracy uses one canonical record per item and endpoint configuration.

Headline values are recomputed from per-item records rather than runtime counters or progress files, which contained audited inconsistencies. A paper-local derivation script recomputes the original leaderboard, paired tests, configuration contrasts, and efficiency values from the hardening matrix and raw DeepSeek records.

\subsection{Statistical definitions}
\label{app:statistics}

For a subset with $k$ key-matching responses among $n$ records, accuracy is $100k/n$. Invalid final answers remain in $n$ and contribute zero to $k$. Wilson score intervals use $z=1.96$ as nominal finite-list descriptions. For two systems evaluated on the same item set, $b$ counts items correct only for the first system and $c$ counts items correct only for the second. The exact two-sided McNemar test treats $b$ as a binomial variate conditional on $b+c$, and Holm--Bonferroni correction is applied across the 15 original-set system pairs. The six original-set native/disabled Overall/Theory/Case tests form a separate Holm family. Final-set tests are post-selection summaries because the native outcomes helped determine inclusion; no selective-inference correction is claimed.

For item $i$, empirical difficulty is
\[
d_i = 1 - \frac{k_i}{6},
\]
where $k_i$ is the number of correct answers among the six native models. The project-specific discrimination diagnostic is
\[
q_i = \tfrac{1}{2}\max(0,r_{i,\mathrm{rest}}) +
      \tfrac{1}{2}\frac{4k_i(6-k_i)}{36},
\]
where $r_{i,\mathrm{rest}}$ is the item-rest point-biserial correlation. With only six model observations per item, this score is a panel-conditional diagnostic, not a stable psychometric item parameter. The paper therefore reports it descriptively and does not claim an item-response-theory calibration.

\section{Secondary Results}
\label{app:robustness}

Invalid-exclusion sensitivity is reported in Table~\ref{tab:invalid-sensitivity}. The top-three order changes from Kimi-K3, GLM-5.3, Qwen3.8-Max under the operational invalid-as-wrong rule to Qwen3.8-Max, Kimi-K3, GLM-5.3 among valid responses. This alternative denominator is not a replacement leaderboard because it conditions on successful gateway completion.

The answer-position audit finds no significant deviation from a uniform predicted A/B/C/D distribution for any main model ($p\geq0.127$). This test does not exclude other annotation artifacts. The structural audit reports longest-option frequencies and substantial chart reuse in the original Case set.

\begin{figure*}[t]
\centering
\tightfig[\textwidth]{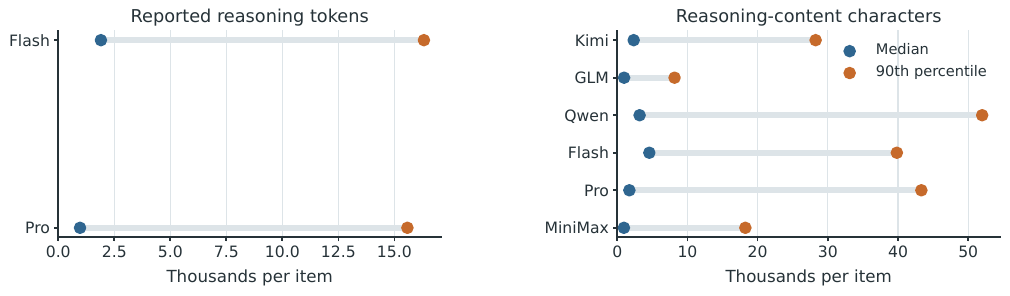}
\caption{Reasoning-effort summaries: median and 90th percentile per item. The token panel includes only the two direct DeepSeek runs with reliable positive token reports. The character panel covers all six systems and treats absent reasoning content as zero; characters remain a route-dependent proxy.}
\label{fig:reasoning-distribution}
\end{figure*}

\subsection{Full aggregate results}
Tables~\ref{tab:macro-scores}, \ref{tab:preselection-leaderboard}, \ref{tab:theory-case}, and \ref{tab:invalid-sensitivity} preserve macro scoring, the original-set reference, task contrasts, and invalid-exclusion sensitivity. Task-macro scores are 1.61--2.47 points lower than final micro scores; category-macro scores are 1.38--2.20 points lower, with unchanged ordering.

\begin{table}[t]
\centering
\caption{Final-set micro and macro scores. Task macro is the equal-weight mean of Theory and Case accuracy; category macro is the equal-weight mean over 25 category accuracies.}
\label{tab:macro-scores}
\scriptsize
\begin{tabular}{lrrr}
\toprule
System & Micro & Task macro & Category macro \\
\midrule
Kimi-K3 & 75.56 & 73.46 & 73.69 \\
GLM-5.3 & 75.08 & 73.17 & 73.38 \\
Qwen3.8-Max & 74.84 & 72.89 & 73.13 \\
DS V4.1 Flash & 72.43 & 70.20 & 70.41 \\
DS V4 Pro & 72.35 & 69.88 & 70.15 \\
MiniMax-M3 & 63.20 & 61.59 & 61.82 \\
\bottomrule
\end{tabular}
\end{table}

\begin{table*}[t]
\centering
\caption{Pre-selection reference results on the original 3,000 items. Intervals are nominal Wilson 95\% intervals and do not account for source/chart clustering.}
\label{tab:preselection-leaderboard}
\small
\begin{tabular}{lrrrr}
\toprule
Endpoint alias & Overall & 95\% CI & Theory & Case \\
\midrule
Kimi-K3 & 79.50 & 78.02--80.91 & 86.40 & 72.60 \\
GLM-5.3 & 79.13 & 77.64--80.55 & 85.00 & 73.27 \\
Qwen3.8-Max & 78.97 & 77.47--80.39 & 85.00 & 72.93 \\
DS V4.1 Flash & 76.97 & 75.43--78.44 & 84.00 & 69.93 \\
DS V4 Pro & 76.83 & 75.29--78.31 & 85.07 & 68.60 \\
MiniMax-M3 & 69.07 & 67.39--70.70 & 72.07 & 66.07 \\
\bottomrule
\end{tabular}
\end{table*}

\begin{table}[t]
\centering
\caption{Theory and Case accuracy and primary gap (percentage points).}
\label{tab:theory-case}
\scriptsize
\begin{tabular}{lrrr}
\toprule
Model & Theory & Case & Gap \\
\midrule
Kimi-K3 & 86.04 & 60.89 & 25.15 \\
GLM-5.3 & 84.59 & 61.75 & 22.84 \\
Qwen3.8-Max & 84.59 & 61.18 & 23.42 \\
DS V4.1 Flash & 83.56 & 56.84 & 26.72 \\
DS V4 Pro & 84.66 & 55.11 & 29.56 \\
MiniMax-M3 & 71.25 & 51.93 & 19.32 \\
\bottomrule
\end{tabular}
\end{table}

\begin{table}[t]
\centering
\caption{Sensitivity to excluding invalid responses from each model's denominator.}
\label{tab:invalid-sensitivity}
\scriptsize
\begin{tabular}{lrrr}
\toprule
Model & Primary & Excl. invalid & Change \\
\midrule
Kimi-K3 & 75.56 & 79.52 & +3.96 \\
GLM-5.3 & 75.08 & 75.26 & +0.18 \\
Qwen3.8-Max & 74.84 & 79.60 & +4.76 \\
DS V4.1 Flash & 72.43 & 72.43 & 0.00 \\
DS V4 Pro & 72.35 & 72.35 & 0.00 \\
MiniMax-M3 & 63.20 & 63.89 & +0.69 \\
\bottomrule
\end{tabular}
\end{table}

\begin{figure}[t]
\centering
\tightfig{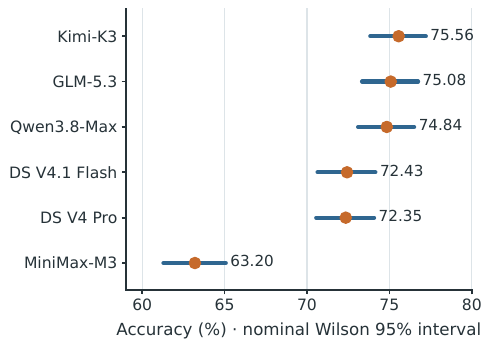}
\caption{Final-set micro accuracy with nominal 95\% Wilson intervals. Values are percentages; intervals are descriptive after outcome-informed selection.}
\label{fig:leaderboard}
\end{figure}
\begin{figure}[t]
\centering
\tightfig{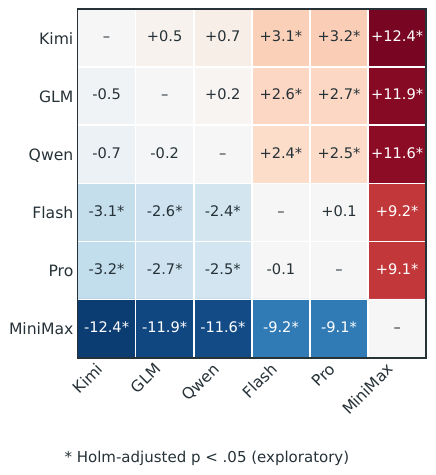}
\caption{Final-set pairwise accuracy differences in percentage points (row minus column). Asterisks mark nominal exact McNemar $p<.05$ after Holm correction across 15 pairs; these post-selection markers are exploratory.}
\label{fig:pairwise}
\end{figure}
Figures~\ref{fig:leaderboard} and \ref{fig:pairwise} give the final-set accuracy visualization and complete pairwise difference matrix. Their nominal intervals and significance markers are descriptive after model-informed selection.

\subsection{Fine-grained analysis}
\label{sec:fine}

\subsubsection{Category structure}

The final benchmark has nearly equal Theory category sizes (102--104 items) and mostly equal Case category sizes (96 items, with 78 Luck Pillars items). Balancing reduces the weight of the largest source category in broad comparisons while retaining categories below the target size.

Across models, Twelve Stages and Nayin are the easiest Theory categories (mean empirical difficulty 0.109 and 0.114), whereas Shensha Basics, Pattern Basics, and Strength and Weakness are the hardest. On Case items, Career and Family Relations are the most difficult (mean difficulty 0.630 and 0.618, with no six-model-all-correct items), followed by Romance; Luck Pillars is the easiest Case category (0.154). These summaries describe agreement with the released gold answers, not the truth of competing interpretive schools.

\begin{table}[t]
\centering
\caption{Most and least difficult categories by six-model empirical difficulty.}
\label{tab:fine-categories}
\scriptsize
\begin{tabular}{llrr}
\toprule
Task & Category & $n$ & Difficulty \\
\midrule
Theory & Shensha Basics & 104 & 0.240 \\
Theory & Pattern Basics & 104 & 0.220 \\
Theory & Strength and Weakness & 104 & 0.210 \\
Theory & Nayin & 104 & 0.114 \\
Theory & Twelve Stages & 104 & 0.109 \\
\midrule
Case & Career & 96 & 0.630 \\
Case & Family Relations & 96 & 0.618 \\
Case & Romance & 96 & 0.573 \\
Case & Personality & 96 & 0.243 \\
Case & Luck Pillars & 78 & 0.154 \\
\bottomrule
\end{tabular}
\end{table}

\begin{figure}[t]
\centering
\tightfig{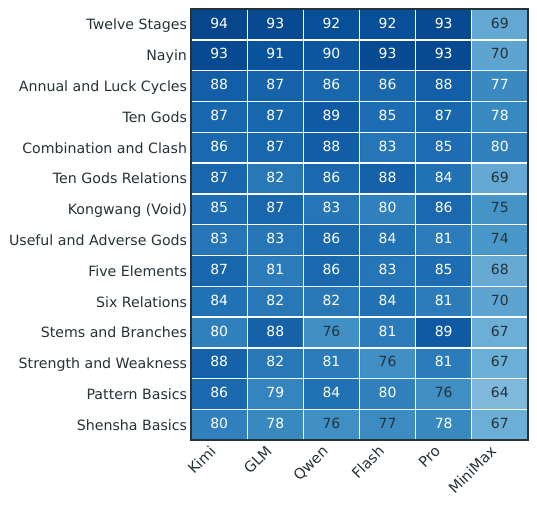}
\caption{Theory-category accuracy across the six models. Vector labels show rounded percentages; full category names follow Table~\ref{tab:taxonomy}.}
\label{fig:theory-heatmap}
\end{figure}

\begin{figure}[t]
\centering
\tightfig{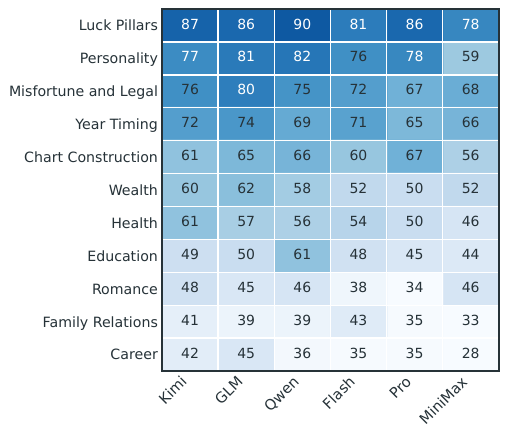}
\caption{Case-category accuracy across the six models. Vector labels show rounded percentages; the larger spread in several categories is consistent with the aggregate Theory-to-Case gap.}
\label{fig:case-heatmap}
\end{figure}

\subsubsection{Difficulty and agreement}

At the item level, 1,118 questions (44.86\%) are answered correctly by all six models and 204 (8.19\%) by none. The distribution is therefore bimodal rather than smoothly difficult. Case items have an all-correct rate of 25.3\% and an all-wrong rate of 13.2\%, compared with 58.8\% and 4.6\% for Theory. The pattern motivates the hardening procedure's focus on concordant denominator mass, but it also warns that a model-informed subset can alter the difficulty spectrum.

\begin{figure}[t]
\centering
\tightfig{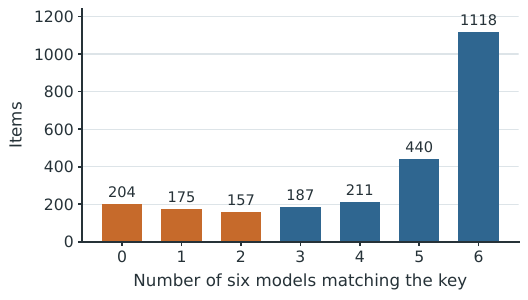}
\caption{Final-set item counts by number of models matching the released key, pooled across Theory and Case (2,492 items).}
\label{fig:difficulty}
\end{figure}

\subsubsection{Answer-position and structural checks}

The target answer distribution is approximately balanced, and chi-square goodness-of-fit tests on model predictions do not reject a uniform A/B/C/D distribution ($p\geq0.127$; Cohen's $w\leq0.048$). No marginal answer-position imbalance was detected; this does not exclude item-conditioned position effects or other annotation artifacts. A separate structural audit finds that the correct option is uniquely the longest in 765/3,000 original items (25.50\%; 284 Theory and 481 Case) and 564/2,492 final items (22.63\%; 278 Theory and 286 Case), as shown in Table~\ref{tab:structural-audit}. These frequencies are structural diagnostics; they alone do not establish an exploitable answer-length heuristic. No item text was altered to change these patterns.

\begin{table}[t]
\centering
\caption{Structural-cue and repeated-chart audit. The cue is present when the
correct option is uniquely the longest. Chart statistics apply to Case items;
``shared items'' counts Case records whose chart occurs at least twice.}
\label{tab:structural-audit}
\scriptsize
\begin{tabular}{llrrrr}
\toprule
Release & Task & $n$ & Cue $n$ (\%) & Charts & Max reuse \\
\midrule
Original & Theory & 1,500 & 284 (18.93) & -- & -- \\
Original & Case & 1,500 & 481 (32.07) & 274 & 33 \\
Original & Total & 3,000 & 765 (25.50) & -- & -- \\
Final & Theory & 1,454 & 278 (19.12) & -- & -- \\
Final & Case & 1,038 & 286 (27.55) & 249 & 25 \\
Final & Total & 2,492 & 564 (22.63) & -- & -- \\
\bottomrule
\end{tabular}
\end{table}

The Case portion also reuses chart contexts. The original 1,500 Case items contain 274 distinct charts, with 1,449 records belonging to a chart used at least twice and a maximum reuse of 33. The final 1,038 Case items contain 249 distinct charts, 974 shared-chart records, and a maximum reuse of 25. These counts motivate treating McNemar and Wilson calculations as fixed-list diagnostics rather than population inference; no cluster bootstrap is claimed.

\subsection{Configuration contrasts and answer transitions}
\label{app:configuration}
\begin{figure}[t]
\centering
\tightfig{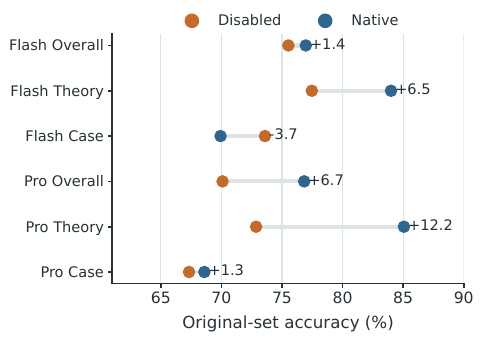}
\caption{Native and disabled-configuration accuracy on the original 3,000 items. Connected dots show paired provider configurations; displayed deltas are native minus disabled.}
\label{fig:ablation}
\end{figure}

Flash has 328 native-key-match/disabled-mismatch original items versus 285 in the reverse direction; Pro has 465 versus 263. These counts explain the net gains without implying a causal failure mechanism. Figure~\ref{fig:ablation} visualizes the original-set contrast, while Table~\ref{tab:ablation-selected} preserves the stronger selected-final diagnostic; the latter conditions on native-arm-informed selection and is not the primary ablation.

\begin{table}[t]
\centering
\caption{Selected-final-set native/disabled diagnostic. Because inclusion uses the native-arm outcomes, these deltas and nominal adjusted $p$-values are post-selection summaries, not confirmatory effects.}
\label{tab:ablation-selected}
\scriptsize
\begin{tabular}{llrrr}
\toprule
Model & Subset & Native & Disabled & $\Delta$ \\
\midrule
Flash & Overall & 72.43 & 71.79 & +0.64 \\
Flash & Theory & 83.56 & 77.03 & +6.53 \\
Flash & Case & 56.84 & 64.45 & -7.61 \\
Pro & Overall & 72.35 & 66.29 & +6.06 \\
Pro & Theory & 84.66 & 72.49 & +12.17 \\
Pro & Case & 55.11 & 57.61 & -2.50 \\
\bottomrule
\end{tabular}
\end{table}

\subsection{Efficiency and failure statistics}
\label{sec:efficiency}

\subsubsection{Reasoning efficiency}

On the original 3,000 items, native Pro is associated with 202 net additional key matches for 14,378,326 reported reasoning tokens, or 71,180 tokens per net match and 1.405 accuracy points per 1,000 additional reasoning tokens per item. Flash has 43 net matches for 15,219,104 tokens, or 353,933 tokens per net match and 0.283 points on the same measure. These unstable descriptive ratios divide by net gain; Flash's overall difference is not detected. They are not monetary costs and are not comparable to mixed-route systems whose reasoning-token fields are incomplete or zero-filled.

On the original set, Flash's mean latency was 28.22 seconds in the native configuration and 1.53 seconds in the disabled configuration. Pro's corresponding means were 91.39 and 2.22 seconds. Native/disabled medians were 10.68/1.43 seconds for Flash and 20.95/2.07 seconds for Pro, giving ratios of medians of 7.44$\times$ and 10.13$\times$.

Request durations include network delays, queueing, and provider execution. The configurations ran at different times, so their latency contrast describes an operational association rather than a causal or compute-only effect.

\begin{table}[H]
\centering
\caption{Token and request-latency efficiency for the paired DeepSeek
configurations on the original 3,000 items. $\Delta C$ is the native-minus-
disabled net key-match count; pp/1k is accuracy-point gain per 1,000
additional reasoning tokens per item. Latency is native/disabled seconds;
the ratio divides the two medians.}
\label{tab:efficiency}
\scriptsize
\setlength{\tabcolsep}{3pt}
\begin{tabular}{lrrrrr}
\toprule
Alias & $\Delta C$ & Tok./$\Delta C$ & pp/1k & Lat. med. N/D & Ratio \\
\midrule
Flash & 43 & 353,933 & 0.283 & 10.68/1.43 & 7.44$\times$ \\
Pro & 202 & 71,180 & 1.405 & 20.95/2.07 & 10.13$\times$ \\
\bottomrule
\end{tabular}
\end{table}

The six V1 native runs show that key-mismatching answers contain more reasoning text than key-matching answers for every system, making length a poor proxy for successful deliberation. On the original DeepSeek runs, Case items consume more output tokens than Theory: 7,535.8 versus 2,614.2 per item for Flash and 8,003.2 versus 1,586.3 for Pro. Latency should not be used as a cross-system speed ranking.

\begin{figure}[t]
\centering
\tightfig{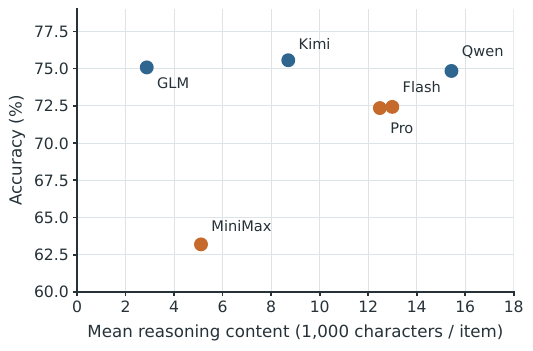}
\caption{Overall accuracy versus mean reasoning-content length over all final-set records, including zeros for absent content. Characters are used because reported reasoning tokens are not comparable across routes. This descriptive cross-system plot is confounded by route and task composition.}
\label{fig:accuracy-reasoning}
\end{figure}

\subsubsection{Invalid responses}

There are 306 invalid responses among the 14,952 six-system final-set evaluations (2.05\%). Qwen3.8-Max has 149, Kimi-K3 124, MiniMax-M3 27, and GLM-5.3 6; both DeepSeek native runs have zero. By recorded source, the micuapi route has 278 invalids in 3,263 records (8.52\%), compared with 15/3,804 for SCNet and 13/2,901 for an early segment whose endpoint is not recorded. The official DeepSeek route has 0/4,984. Of the invalid records, 292 contain positive-length reasoning content and 14 Kimi/micuapi records do not; finish reasons are blank for 266, \code{error} for seven, and \code{stop} for 33, with no \code{length} records. These observations support an interface or route diagnosis, but do not prove that an underlying model is intrinsically incapable of answering an item.

\begin{table}[t]
\centering
\caption{Invalid responses in the six-model main evaluation.}
\label{tab:invalids}
\scriptsize
\begin{tabular}{lrr}
\toprule
Model & Invalid $n$ & Invalid (\%) \\
\midrule
Qwen3.8-Max & 149 & 5.98 \\
Kimi-K3 & 124 & 4.98 \\
MiniMax-M3 & 27 & 1.08 \\
GLM-5.3 & 6 & 0.24 \\
DS V4.1 Flash & 0 & 0.00 \\
DS V4 Pro & 0 & 0.00 \\
\midrule
Total & 306 & 2.05 \\
\bottomrule
\end{tabular}
\end{table}

\begin{figure}[t]
\centering
\tightfig{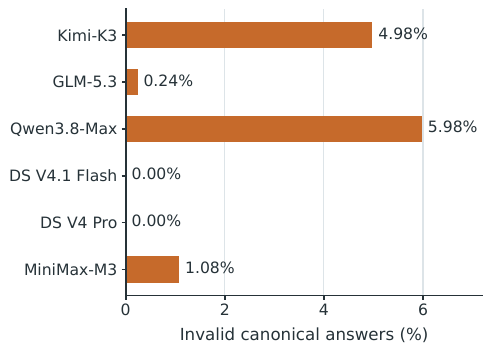}
\caption{Invalid response rates on the final set. Empty answers are counted as wrong in the primary accuracy metric.}
\label{fig:invalid}
\end{figure}

\subsubsection{Failure concentration}

The 204 final-set items unmatched by all six answer vectors are concentrated in Career, Romance, Chart Construction, and Education Case categories. On the original set, the paired configurations create bidirectional reversals: 285 Flash items and 263 Pro items change from a disabled-configuration key match to a native-configuration mismatch. We do not infer a causal failure mechanism. Without human-coded examples, these are failure statistics and candidate strata for expert adjudication, not a qualitative error analysis.

\subsubsection{Source and latency sensitivity}

Routes covary with task order. The micuapi segment is mostly Case; early and SCNet segments are mostly Theory. Accuracy differences between segments thus combine route and item-composition effects. We use source primarily to diagnose invalid responses and restrict efficiency interpretation to clean direct comparisons.

\section{Reproducibility and Release}
\label{app:reproducibility}

The canonical final analysis contains eight complete runs (six native and two disabled), yielding $8\times2{,}492=19{,}936$ retained-item evaluations. The archived raw response files cover the original 3,000-item run for each of the same eight configurations. The paper package retains CSV snapshots for all reported tables and PDF/PNG copies of deterministic figures. The original analysis scripts remain the source of truth; paper-local derivation helpers accept their root path rather than embedding a machine-specific location.

The release history required to interpret the paper is:
\begin{enumerate}
    \item archived original JSONL files with 1,500 Theory and 1,500 Case items;
    \item per-item main and ablation responses;
    \item item-metric and hardening analyses;
    \item the fixed-seed final subset builder;
    \item final JSONL files and the 508-row deletion manifest.
\end{enumerate}

BaZi2500 is the public benchmark release under CC BY-NC 4.0; resource links are provided in the main-text disclosures.

\subsection{Responsible use and ethics}
This artifact evaluates culturally situated textual and rule-based QA. It does not establish or endorse real-world predictive validity and should not inform medical, legal, financial, employment, relationship, or other consequential decisions. Historical, school-specific, and culturally contested claims may appear in the source representation; model performance does not adjudicate them.

The study evaluates archived textual QA artifacts and recruited no participants or interventions. The authors have manually reviewed and addressed privacy in the public data.

\subsection{Data availability and author disclosures}
The project preserves original and final releases, the deletion manifest, analysis tables, and deterministic figures. BaZi2500 is publicly released under CC BY-NC 4.0; the authors have confirmed that its classical and traditional source materials have no modern copyright restrictions and may be redistributed in this form. Author names and affiliations are provided on the title page. Contact: \href{mailto:lijiulin@18trees.com}{lijiulin@18trees.com}. This work was supported by Beijing Liuyi Guanhua Technology Co., Ltd. The authors declare no competing interests.

Language models assisted upstream source gating, extraction, drafting, verification, and deduplication and were subjects of evaluation. An agent-assisted workflow assembled this manuscript from project evidence. No model acted as an independent human expert validator; numerical results come from deterministic analysis outputs checked against source artifacts.

\end{document}